\documentclass{article}

\usepackage[final,nonatbib]{nips_2017}

\usepackage{graphicx}

\usepackage{algorithm}
\usepackage{algpseudocode}

\usepackage[utf8]{inputenc} 
\usepackage[T1]{fontenc}    
\usepackage{hyperref}       
\usepackage{url}            
\usepackage{booktabs}       
\usepackage{amsfonts}       
\usepackage{nicefrac}       
\usepackage{microtype}      
\usepackage[misc,geometry]{ifsym} 
\usepackage{amsmath}

\DeclareMathOperator*{\argmax}{argmax}

\usepackage{color}

\usepackage{authblk}

\title{Thinking Fast and Slow\\ with Deep Learning and Tree Search}

\author[1, \Letter]{\textbf{Thomas Anthony}}
\author[1]{\textbf{Zheng Tian}}
\author[1,2]{\textbf{David Barber}}
\affil[1]{University College London}
\affil[2]{Alan Turing Institute}
\affil[ \Letter]{\url{thomas.anthony.14@ucl.ac.uk}}

\begin{document}

\maketitle

\begin{abstract}

Sequential decision making problems, such as structured prediction, robotic control, and game playing, require a combination of planning policies and generalisation of those plans. In this paper, we present Expert Iteration (\textsc{ExIt}), a novel reinforcement learning algorithm which decomposes the problem into separate planning and generalisation tasks. Planning new policies is performed by tree search, while a deep neural network generalises those plans. Subsequently, tree search is improved by using the neural network policy to guide search, increasing the strength of new plans. In contrast, standard deep Reinforcement Learning algorithms rely on a neural network not only to generalise plans, but to discover them too. We show that \textsc{ExIt} outperforms \textsc{REINFORCE} for training a neural network to play the board game Hex, and our final tree search agent, trained tabula rasa, defeats \textsc{MoHex} 1.0, the most recent Olympiad Champion player to be publicly released.
\end{abstract}

\section{Introduction}
\label{intro}

According to dual-process theory \cite{evans1984heuristic, kahneman2003maps}, human reasoning consists of two different kinds of thinking. \textit{System 1} is a fast, unconscious and automatic mode of thought, also known as \textit{intuition} or \textit{heuristic process}. \textit{System 2}, an evolutionarily recent process unique to humans, is a slow, conscious, explicit and rule-based mode of \textit{reasoning}. 

When learning to complete a challenging planning task, such as playing a board game, humans exploit both processes: strong intuitions allow for more effective analytic reasoning by rapidly selecting interesting lines of play for consideration. Repeated deep study gradually improves intuitions. Stronger intuitions feedback to stronger analysis, creating a closed learning loop. In other words, humans learn by \textit{thinking fast and slow}. 

In deep Reinforcement Learning (RL) algorithms such as \textsc{REINFORCE} \cite{williams1992simple} and DQN \cite{mnih2015human}, neural networks make action selections with no lookahead; this is analogous to System 1. Unlike human intuition, their training does not benefit from a `System 2' to suggest strong policies. In this paper, we present Expert Iteration (\textsc{ExIt}), which uses a Tree Search as an analogue of System 2; this assists the training of the neural network. In turn, the neural network is used to improve the performance of the tree search by providing fast `intuitions' to guide search.

At a low level, \textsc{ExIt} can be viewed as an extension of Imitation Learning (IL) methods to domains where the best known experts are unable to achieve satisfactory performance. In IL an \textit{apprentice} is trained to imitate the behaviour of an \textit{expert} policy. Within \textsc{ExIt}, we iteratively re-solve the IL problem. Between each iteration, we perform an expert improvement step, where we bootstrap the (fast) apprentice policy to increase the performance of the (comparatively slow) expert.

Typically, the apprentice is implemented as a deep neural network, and the expert by a tree search algorithm. Expert improvement can be achieved either by using the apprentice as an initial bias in the search direction, or to assist in quickly estimating the value of states encountered in the search tree, or both.

We proceed as follows: in section \ref{prelims}, we cover some preliminaries. Section \ref{general} describes the general form of the Expert Iteration algorithm, and discusses the roles performed by expert and apprentice. 

Sections \ref{ilhex} and \ref{ExItHex} dive into the implementation details of the Imitation Learning and expert improvement steps of \textsc{ExIt} for the board game Hex. The performance of the resultant \textsc{ExIt} algorithm is reported in section \ref{results}. Sections \ref{related} and \ref{conclusion} discuss our findings and relate the algorithm to previous works.

\section{Preliminaries}
\label{prelims}
\subsection{Markov Decision Processes}

We consider sequential decision making in a Markov Decision Process (MDP). At each timestep $t$, an agent observes a state $s_t$ and chooses an action $a_t$ to take. In a terminal state $s_T$, an episodic reward $R$ is observed, which we intend to maximise.\footnote{This reward may be decomposed as a sum of intermediate rewards (i.e. $R = \sum_{t=0}^{T} r_t$)} We can easily extend to two-player, perfect information, zero-sum games by learning policies for both players simultaneously, which aim to maximise the reward for the respective player.

We call a distribution over the actions $a$ available in state $s$ a \textit{policy}, and denote it $\pi(a|s)$. The value function $V^\pi(s)$ is the mean reward from following $\pi$ starting in state $s$. By $Q^\pi(s,a)$ we mean the expected reward from taking action $a$ in state $s$, and following policy $\pi$ thereafter.

\subsection{Imitation Learning}

In Imitation Learning (IL), we attempt to solve the MDP by mimicking an \textit{expert} policy $\pi^{*}$ that has been provided. Experts can arise from observing humans completing a task, or, in the context of structured prediction, calculated from labelled training data. The policy we learn through this mimicry is referred to as the \textit{apprentice} policy.

We create a dataset of states of expert play, along with some target data drawn from the expert, which we attempt to predict. Several choices of target data have been used. The simplest approach is to ask the expert to name an optimal move $\pi^{*}(a|s)$ \cite{Ross2010reduction}. Once we can predict expert moves, we can take the action we think the expert would have most probably taken. Another approach is to estimate the action-value function $Q^{\pi^{*}}(s,a)$. We can then predict that function, and act greedily with respect to it. In contrast to direct action prediction, this target is cost-sensitive, meaning the apprentice can trade-off prediction errors against how costly they are \cite{hal2009search}. 

\section{Expert iteration}
\label{general}

Compared to IL techniques, Expert Iteration (\textsc{ExIt}) is enriched by an expert improvement step. Improving the expert player and then resolving the Imitation Learning problem allows us to exploit the fast convergence properties of Imitation Learning even in contexts where no strong player was originally known, including when learning tabula rasa. Previously, to solve such problems, researchers have fallen back on RL algorithms that often suffer from slow convergence, and high variance, and can struggle with local minima.

At each iteration $i$, the algorithm proceeds as follows: we create a set $\mathcal{S}_i$ of game states by self play of the apprentice $\hat\pi_{i-1}$. In each of these states, we use our expert to calculate an Imitation Learning target at $s$ (e.g. the expert's action $\pi^{*}_{i-1}(a|s)$); the state-target pairs (e.g. $(s,\pi^{*}_{i-1}(a|s))$) form our dataset $\mathcal{D}_i$ . We train a new apprentice $\hat\pi_i$ on $\mathcal{D}_i$ (Imitation Learning). Then, we use our new apprentice to update our expert $\pi^{*}_i = \pi^{*}(a|s;\hat\pi_i)$ (expert improvement). See Algorithm \ref{alg:exit} for pseudo-code.

The expert policy is calculated using a tree search algorithm. By using the apprentice policy to direct search effort towards promising moves, or by evaluating states encountered during search more quickly and accurately, we can help the expert find stronger policies. In other words, we bootstrap the knowledge acquired by Imitation Learning back into the planning algorithm.

The Imitation Learning step is analogous to a human improving their intuition for the task by studying example problems, while the expert improvement step is analogous to a human using their improved intuition to guide future analysis.

\begin{algorithm}[H]
	\caption{Expert Iteration}
	\label{alg:exit}
	\begin{algorithmic}[1]
		\State $\hat{\pi}_0$ = initial\_policy()
		\State $\pi^*_0$ = build\_expert($\hat{\pi}_0$)
		\For{i = 1; i $\leq$ max\_iterations; i++}
		\State $S_i$ = sample\_self\_play($\hat{\pi}_{i-1}$)
		\State $D_i$ = $\{ (s$, imitation\_learning\_target($\pi^*_{i-1}(s))) | s \in S_i \}$
		\State $\hat{\pi}_i$ = train\_policy($D_i$)
		\State $\pi^*_i$ = build\_expert($\hat{\pi}_i$)
		\EndFor
	\end{algorithmic}
\end{algorithm}

\subsection{Choice of expert and apprentice}

The learning rate of \textsc{ExIt} is controlled by two factors: the size of the performance gap between the apprentice policy and the improved expert, and how close the performance of the new apprentice is to the performance of the expert it learns from. The former induces an upper bound on the new apprentice's performance at each iteration, while the latter describes how closely we approach that upper bound. The choice of both expert and apprentice can have a significant impact on both these factors, so must be considered together.

The role of the expert is to perform exploration, and thereby to accurately determine strong move sequences, from a single position. The role of the apprentice is to generalise the policies that the expert discovers across the whole state space, and to provide rapid access to that strong policy for bootstrapping in future searches. 

The canonical choice of expert is a tree search algorithm. Search considers the exact dynamics of the game tree local to the state under consideration. This is analogous to the lookahead human games players engage in when planning their moves. The apprentice policy can be used to bias search towards promising moves, aid node evaluation, or both. By employing search, we can find strong move sequences potentially far away from the apprentice policy, accelerating learning in complex scenarios. Possible tree search algorithms include Monte Carlo Tree Search \cite{kocsis2006bandit}, $\alpha$-$\beta$ Search, and greedy search \cite{hal2009search}.

The canonical apprentice is a deep neural network parametrisation of the policy. Such deep networks are known to be able to efficiently generalise across large state spaces, and they can be evaluated rapidly on a GPU. The precise parametrisation of the apprentice should also be informed by what data would be useful for the expert. For example, if state value approximations are required, the policy might be expressed implicitly through a $Q$ function, as this can accelerate lookup.

\subsection{Distributed Expert Iteration}

Because our tree search is orders of magnitude slower than the evaluations made during training of the neural network, \textsc{ExIt} spends the majority of run time creating datasets of expert moves. Creating these datasets is an embarassingly parallel task, and the plans made can be summarised by a vector measuring well under 1KB. This means that \textsc{ExIt} can be trivially parallelised across distributed architectures, even with very low bandwidth.

\subsection{Online expert iteration}

In each step of \textsc{ExIt}, Imitation Learning is restarted from scratch. This throws away our entire dataset. Since creating datasets is computationally intensive this can add substantially to algorithm run time. 

The online version of \textsc{ExIt} mitigates this by aggregating all datasets generated so far at each iteration. In other words, instead of training $\hat{\pi}_i$ on $\mathcal{D}_i$, we train it on $\mathcal{D} = \cup_{j\le i} \mathcal{D}_j$. Such dataset aggregation is similar to the \textsc{DAgger} algorithm \cite{Ross2010reduction}. Indeed, removing the expert improvement step from online \textsc{ExIt} reduces it to \textsc{DAgger}.

Dataset aggregation in online \textsc{ExIt} allows us to request fewer move choices from the expert at each iteration, while still maintaining a large dataset. By increasing the frequency at which improvements can be made, the apprentice in online \textsc{ExIt} can generalise the expert moves sooner, and hence the expert improves sooner also, which results in higher quality play appearing in the dataset.

\section{Imitation Learning in the game Hex}
\label{ilhex}

We now describe the implementation of \textsc{ExIt} for the board game Hex. In this section, we develop the techniques for our Imitation Learning step, and test them for Imitation Learning of Monte Carlo Tree Search (MCTS). We use this test because our intended expert is a version of Neural-MCTS, which will be described in section \ref{ExItHex}.

\subsection{Preliminaries} 
\subsubsection*{Hex}

\begin{figure}[h!]
	\begin{center}
		\includegraphics[scale=0.3]{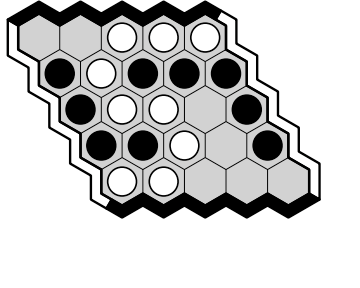}\\
		\vspace*{-8mm}
		\caption{A $5\times5$ Hex game, won by white. Figure from Huang et al. \cite{huang2013mohex}.}
		\label{hex}
	\end{center}
\end{figure}

Hex is a two-player connection-based game played on an $n \times n$ hexagonal grid. The players, denoted by colours black and white, alternate placing stones of their colour in empty cells. The black player wins if there is a sequence of adjacent black stones connecting the North edge of the board to the South edge. White wins if they achieve a sequence of adjacent white stones running from the West edge to the East edge. (See figure \ref{hex}).

Hex requires complex strategy, making it challenging for deep Reinforcement Learning algorithms; its large action set and connection-based rules means it shares similar challenges for AI to Go. However, games can be simulated efficiently because the win condition is mutually exclusive (e.g. if black has a winning path, white cannot have one), its rules are simple, and permutations of move order are irrelevant to the outcome of a game. These properties make it an ideal test-bed for Reinforcement Learning. All our experiments are on a $9\times9$ board size.

\subsubsection*{Monte Carlo Tree Search}

Monte Carlo Tree Search (MCTS) is an any-time best-first tree-search algorithm. It uses repeated game simulations to estimate the value of states, and expands the tree further in more promising lines. When all simulations are complete, the most explored move is taken. It is used by the leading algorithms in the AAAI general game-playing competition \cite{genesereth2005general}. As such, it is the best known algorithm for general game-playing without a long RL training procedure.

Each simulation consists of two parts. First, a \textit{tree phase}, where the tree is traversed by taking actions according to a \textit{tree policy}. Second, a \textit{rollout phase}, where some \textit{default policy} is followed until the simulation reaches a terminal game state. The result returned by this simulation can then be used to update estimates of the value of each node traversed in the tree during the first phase.

Each node of the search tree corresponds to a possible state $s$ in the game. The root node corresponds to the current state, its children correspond to the states resulting from a single move from the current state, etc. The edge from state $s_1$ to $s_2$ represents the action $a$ taken in $s_1$ to reach $s_2$, and is identified by the pair $(s_1,a)$.

At each node we store $n(s)$, the number of iterations in which the node has been visited so far. Each edge stores both $n(s,a)$, the number of times it has been traversed, and $r(s,a)$ the sum of all rewards obtained in simulations that passed through the edge. The tree policy depends on these statistics. The most commonly used tree policy is to act greedily with respect to the upper confidence bounds for trees formula \cite{kocsis2006bandit}:  

\begin{equation} \mathrm{UCT}(s,a) = \frac{r(s,a)}{n(s,a)} + c_b \sqrt{\frac{\log{n(s)}}{n(s,a)}}\end{equation}

When an action $a$ in a state $s_L$ is chosen that takes us to a position $s'$ not yet in the search tree, the rollout phase begins. In the absence of domain-specific information, the default policy used is simply to choose actions uniformly from those available.

To build up the search tree, when the simulation moves from the tree phase to the rollout phase, we perform an expansion, adding $s'$ to the tree as a child of $s_L$.\footnote{Sometimes multiple nodes are added to the tree per iteration, adding children to $s'$ also. Conversely, sometimes an \textit{expansion threshold} is used, so $s_L$ is only expanded after multiple visits.} Once a rollout is complete, the reward signal is propagated through the tree (a \textit{backup}), with each node and edge updating statistics for visit counts $n(s)$, $n(s,a)$ and total returns $r(s,a)$.

In this work, all MCTS agents use 10,000 simulations per move, unless stated otherwise. All use a uniform default policy. We also use RAVE. Full details are in the appendix. \cite{gelly2007combining}.

\subsection{Imitation Learning from Monte Carlo Tree Search}
\label{ilmcts}

In this section, we train a standard convolutional neural network\footnote{Our network architecture is described in the appendix. We use Adam \cite{kingma2014adam} as our optimiser.} to imitate an MCTS expert. Guo et al. \cite{guo2014deep} used a similar set up on Atari games. However, their results showed that the performance of the learned neural network fell well short of the MCTS expert, even with a large dataset of 800,000 MCTS moves. Our methodology described here improves on this performance.

\subsubsection*{Learning Targets}

In Guo et al. \cite{guo2014deep}, the learning target used was simply the move chosen by MCTS. We refer to this as \textit{chosen-action targets} (CAT), and optimise the Kullback–Leibler divergence between the output distribution of the network and this target. So the loss at position $s$ is given by the formula:

$$\mathcal{L}_{\mathrm{CAT}} = -\log[\pi(a^*|s)]$$

where $a^{*} = \argmax_a(n(s,a))$ is the move selected by MCTS.

We propose an alternative target, which we call \textit{tree-policy targets} (TPT). The tree policy target is the average tree policy of the MCTS at the root. In other words, we try to match the network output to the distribution over actions given by $n(s,a)/n(s)$ where $s$ is the position we are scoring (so $n(s)= 10,000$ in our experiments). This gives the loss:

$$\mathcal{L}_{\mathrm{TPT}} = -\sum_a \frac{n(s,a)}{n(s)}\log[\pi(a|s)]$$

Unlike CAT, TPT is cost-sensitive: when MCTS is less certain between two moves (because they are of similar strength), TPT penalises misclassifications less severely. Cost-sensitivity is a desirable property for an imitation learning target, as it induces the IL agent to trade off accuracy on less important decisions for greater accuracy on critical decisions.

In \textsc{ExIt}, there is additional motivation for such cost-sensitive targets, as our networks will be used to bias future searches. Accurate evaluations of the relative strength of actions never made by the current expert are still important, since future experts will use the evaluations of all available moves to guide their search.

\subsubsection*{Sampling the position set}

Correlations between the states in our dataset may reduce the effective dataset size, harming learning. Therefore, we construct all our datasets to consist of uncorrelated positions sampled using an exploration policy. To do this, we play multiple games with an exploration policy, and select a single state from each game, as in Silver et al. \cite{silver2016mastering}. For the initial dataset, the exploration policy is MCTS, with the number of iterations reduced to 1,000 to reduce computation time and encourage a wider distribution of positions.

We then follow the \textsc{DAgger} procedure, expanding our dataset by using the most recent apprentice policy to sample 100,000 more positions, again sampling one position per game to ensure that there were no correlations in the dataset. This has two advantages over sampling more positions in the same way: firstly, selecting positions with the apprentice is faster, and secondly, doing so results in positions closer to the distribution that the apprentice network visits at test time.

\subsection{Results of Imitation Learning}

Based on our initial dataset of 100,000 MCTS moves, CAT and TPT have similar performance in the task of predicting the move selected by MCTS, with average top-1 prediction errors of $47.0\%$ and $47.7\%$, and top-3 prediction errors of $65.4\%$ and $65.7\%$, respectively.
	
However, despite the very similar prediction errors, the TPT network is $50\pm13$ Elo stronger than the CAT network, suggesting that the cost-awareness of TPT indeed gives a performance improvement. \footnote{When testing network performance, we greedily select the most likely move, because CAT and TPT may otherwise induce different temperatures in the trained networks' policies.}

We continued training the TPT network with the \textsc{DAgger} algorithm, iteratively creating 3 more batches of 100,000 moves. This additional data resulted in an improvement of $120$ Elo over the first TPT network. Our final \textsc{DAgger} TPT network achieved similar performance to the MCTS it was trained to emulate, winning just over half of games played between them ($87/162$).

\section{Expert Improvement in Hex}
\label{ExItHex}

We now have an Imitation Learning procedure that can train a strong apprentice network from MCTS. In this section, we describe our Neural-MCTS (N-MCTS) algorithms, which use such apprentice networks to improve search quality. 

\subsection{Using the Policy Network}

Because the apprentice network has effectively generalised our policy, it gives us fast evaluations of action plausibility at the start of search. As search progresses, we discover improvements on this apprentice policy, just as human players can correct inaccurate intuitions through lookahead.

We use our neural network policy to bias the MCTS tree policy towards moves we believe to be stronger. When a node is expanded, we evaluate the apprentice policy $\hat{\pi}$ at that state, and store it. We modify the UCT formula by adding a bonus proportional to $\hat{\pi}(a|s)$:

$$\mathrm{UCT}_{\mathrm{P-NN}}(s,a) = \mathrm{UCT}(s,a) + w_a\frac{\hat{\pi}(a|s)}{n(s,a)+1}$$

Where $w_a$ weights the neural network against the simulations. This formula is adapted from one found in Gelly \& Silver \cite{gelly2007combining}. Tuning of hyperparameters found that $w_a = 100$ was a good choice for this parameter, which is close to the average number of simulations per action at the root when using 10,000 iterations in the MCTS. Since this policy was trained using 10,000 iterations too, we would expect that the optimal weight should be close to this average.

The TPT network's final layer uses a softmax output. Because there is no reason to suppose that the optimal bonus in the UCT formula should be linear in the TPT policy probability, we view the temperature of the TPT network's output layer as a hyperparameter for the N-MCTS and tune it to maximise the performance of the N-MCTS.

When using the strongest TPT network from section \ref{ilhex}, N-MCTS using a policy network significantly outperforms our baseline MCTS, winning $97\%$ of games. The neural network evaluations cause a two times slowdown in search. For comparison, a doubling of the number of iterations of the vanilla MCTS results in a win rate of $56\%$.

\subsection{Using a Value Network}
\label{vmcts}

Strong value networks have been shown to be able to substantially improve the performance of MCTS \cite{silver2016mastering}. Whereas a policy network allows us to narrow the search, value networks act to reduce the required search depth compared to using inaccurate rollout-based value estimation.

However, our imitation learning procedure only learns a policy, not a value function. Monte Carlo estimates of $V^{\pi^*}(s)$ could be used to train a value function, but to train a value function without severe overfitting requires more than $10^5$ independent samples. Playing this many expert games is well beyond our computation resources, so instead we approximate $V^{\pi^*}(s)$ with the value function of the apprentice, $V^{\hat{\pi}}(s)$, for which Monte Carlo estimates are cheap to produce.

To train the value network, we use a KL loss between $V(s)$ and the sampled (binary) result $z$: $$\mathcal{L}_{\mathrm{V}} = -z\log[V(s)] - (1-z)\log[1-V(s)]$$

To accelerate the tree search and regularise value prediction, we used a multitask network with separate output heads for the apprentice policy and value prediction, and sum the losses $\mathcal{L}_{\mathrm{V}}$ and $\mathcal{L}_{\mathrm{TPT}}$.

To use such a value network in the expert, whenever a leaf $s_L$ is expanded, we estimate $V(s)$. This is backed up through the tree to the root in the same way as rollout results are: each edge stores the average of all evaluations made in simulations passing through it. In the tree policy, the value is estimated as a weighted average of the network estimate and the rollout estimate.\footnote{This is the same as the method used in Silver et al. \cite{silver2016mastering}}

\section{Experiments}
\label{results}

\subsection{Comparison of Batch and Online \textsc{ExIt} to REINFORCE}

We compare \textsc{ExIt} to the policy gradient algorithm found in Silver et al. \cite{silver2016mastering}, which achieved state-of-the-art performance for a neural network player in the related board game Go. In Silver et al. \cite{silver2016mastering}, the algorithm was initialised by a network trained to predict human expert moves from a corpus of 30 million positions, and then \textsc{REINFORCE} \cite{williams1992simple} was used. We initialise with the best network from section \ref{ilhex}. Such a scheme, Imitation Learning initialisation followed by Reinforcement Learning improvement, is a common approach when known experts are not sufficiently strong.

In our batch \textsc{ExIt}, we perform 3 training iterations, each time creating a dataset of 243,000 moves. 

In online \textsc{ExIt}, as the dataset grows, the supervised learning step takes longer, and in a na\"ive implementation would come to dominate run-time. We test two forms of online \textsc{ExIt} that avoid this. In the first, we create 24,300 moves each iteration, and train on a buffer of the most recent 243,000 expert moves. In the second, we use all our data in training, and expand the size of the dataset by $10\%$ each iteration. 

For this experiment we did not use any value networks, so that network architectures between the policy gradient and \textsc{ExIt} are identical. All policy networks are warm-started to the best network from section \ref{ilhex}.

\begin{figure}[h!]
	\begin{center}
		\includegraphics[scale=0.5]{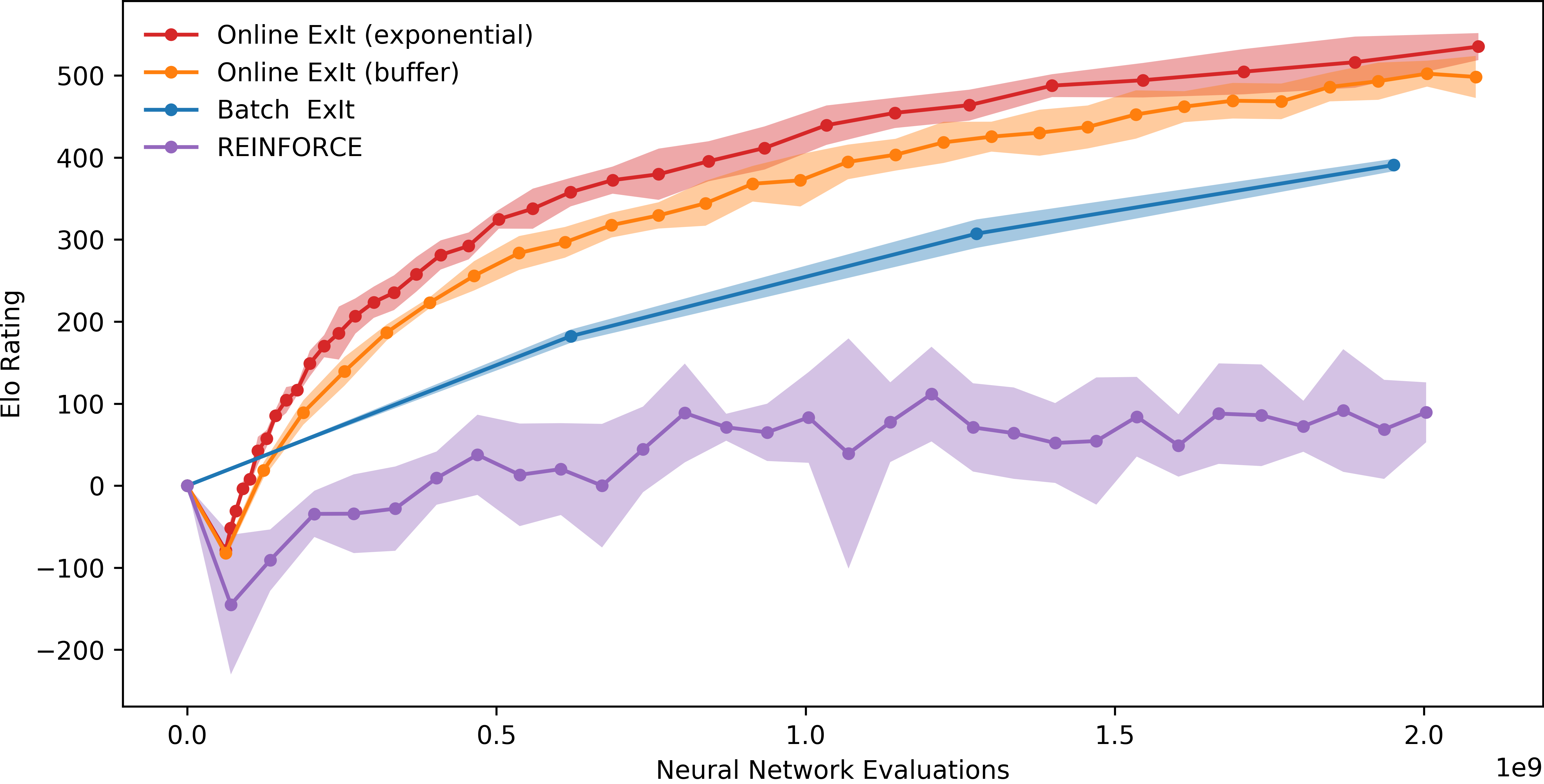}\\
		\vspace*{-1mm}
		\caption{Elo ratings of policy gradient network and \textsc{ExIt} networks through training. Values are the average of 5 training runs, shaded areas represent $90\%$ confidence intervals. Time is measured by number of neural network evaluations made. Elo calculated with BayesElo \cite{bayeselo}}.
		\label{short_exp}
	\end{center}
\end{figure}

As can be seen in figure \ref{short_exp}, compared to \textsc{REINFORCE}, \textsc{ExIt} learns stronger policies faster. \textsc{ExIt} also shows no sign of instability: the policy improves consistently each iteration and there is little variation in the performance between each training run. Separating the tree search from the generalisation has ensured that plans don't overfit to a current opponent, because the tree search considers multiple possible responses to the moves it recommends.

Online expert iteration substantially outperforms the batch mode, as expected. Compared to the `buffer' version, the `exponential dataset' version appears to be marginally stronger, suggesting that retaining a larger dataset is useful. 

\subsection{Comparison of Value and Policy \textsc{ExIt}}

With sufficiently large datasets, a value network can be learnt to improve the expert further, as discussed in section \ref{vmcts}. We ran asynchronous distributed online \textsc{ExIt} using only a policy network until our datasets contained $\sim550,000$ positions. We then used our most recent apprentice to add a Monte Carlo value estimate from each of the positions in our dataset, and trained a combined policy and value apprentice, giving a substantial improvement in the quality of expert play. 

\begin{figure}[h!]
	\begin{center}
		\includegraphics[scale=0.5]{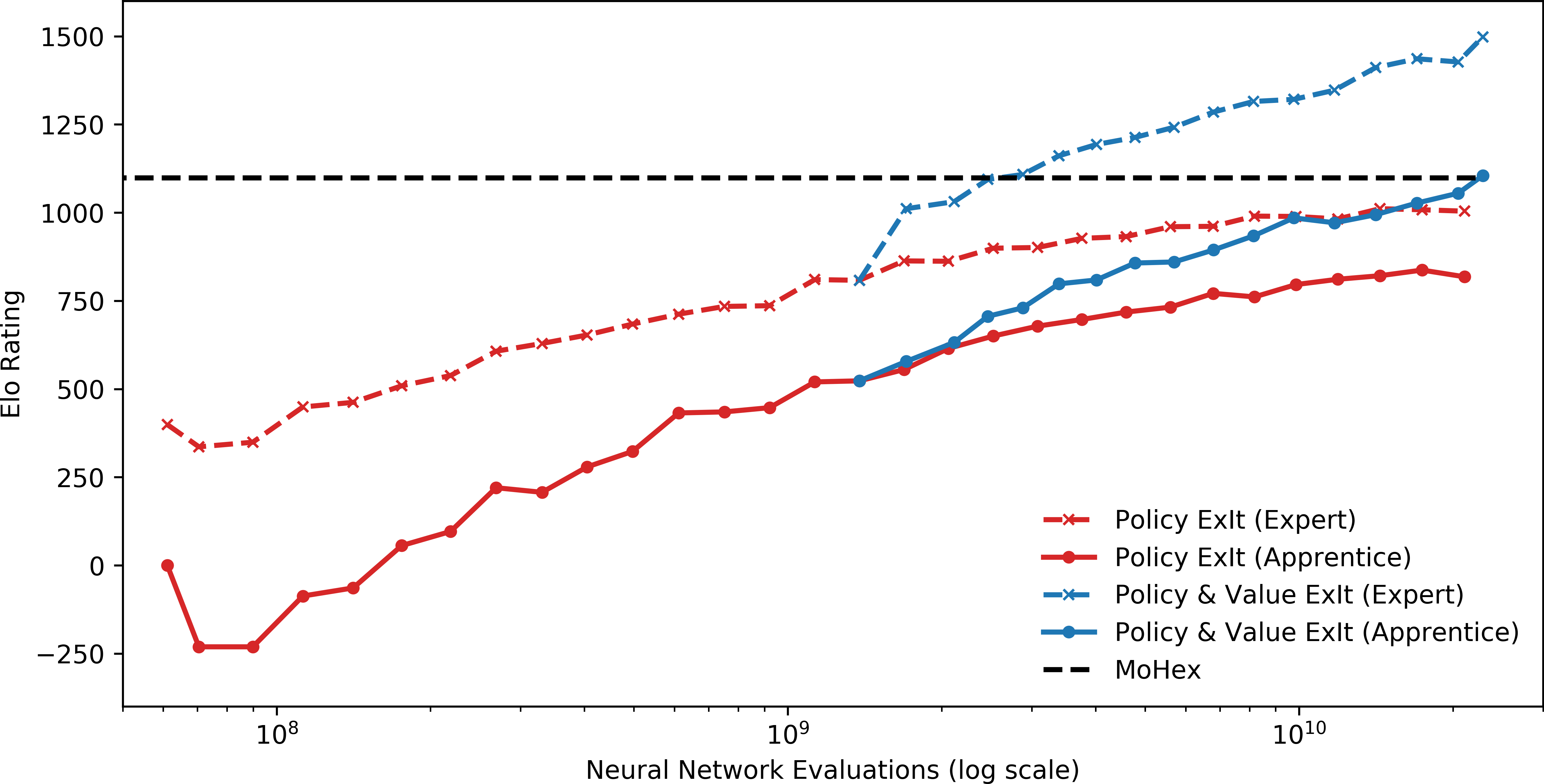}\\
		\vspace*{-1mm}
		\caption{Apprentices and experts in distributed online \textsc{ExIt}, with and without neural network value estimation. \textsc{MoHex}'s rating (10,000 iterations per move) is shown by the black dashed line.}
		\label{long_exp}
	\end{center}
\end{figure}

We then ran \textsc{ExIt} with a combined value-and-policy network, creating another $\sim7,400,000$ move choices. For comparison, we continued the training run without using value estimation for equal time. Our results are shown in figure \ref{long_exp}, which shows that value-and-policy-\textsc{ExIt} significantly outperforms policy-only-\textsc{ExIt}. In particular, the improved plans from the better expert quickly manifest in a stronger apprentice.

We can also clearly see the importance of expert improvement, with later apprentices comfortably outperforming experts from earlier in training.

\subsection{Performance Against \textsc{MoHex}}

Versions of \textsc{MoHex} have won every Computer Games Olympiad Hex tournament since 2009. \textsc{MoHex} is a highly optimised algorithm, utilising a complex, hand-made theorem-proving algorithm which calculates provably suboptimal moves, to be pruned from search, and an improved rollout policy; it also optionally uses a specialised end-game solver, particularly powerful for small board sizes. In contrast, our algorithm learns tabula rasa, without game-specific knowledge beside the rules of the game. Here we compare to the most recent available version, \textsc{MoHex} 1.0 \cite{arneson2010hex}.

To fairly compare \textsc{MoHex} to our experts with equal wall-clock times is difficult, as the relative speeds of the algorithms are hardware dependent: \textsc{MoHex}'s theorem prover makes heavy use of the CPU, whereas for our experts, the GPU is the bottleneck. On our machine \textsc{MoHex} is approximately $50\%$ faster.\footnote{This machine has an Intel Xeon E5-1620 and nVidia Titan X (Maxwell), our tree search takes 0.3 seconds for 10,000 iterations, while \textsc{MoHex} takes 0.2 seconds for 10,000 iterations, with multithreading.}

\textsc{ExIt} (with 10,000 iterations) won $75.3\%$ of games against 10,000 iteration-\textsc{MoHex} on default settings, $59.3\%$ against 100,000 iteration-\textsc{MoHex}, and $55.6\%$ against 4 second per move-\textsc{Mohex} (with parallel solver switched on), which are over six times slower than our searcher. We include some sample games from the match between 100,000 iteration \textsc{MoHex} and \textsc{ExIt} in the appendix. This result is particularly remarkable because the training curves in figure \ref{long_exp} do not suggest that the algorithm has reached convergence.

\section{Related work}
\label{related}

\textsc{ExIt} has several connections to existing RL algorithms, resulting from different choices of expert class. For example, we can recover a version of Policy Iteration \cite{ross2014reinforcement} by using Monte Carlo Search as our expert; in this case it is easy to see that Monte Carlo Tree Search gives stronger plans than Monte Carlo Search.

Previous works have also attempted to achieve Imitation Learning that outperforms the original expert. Silver et al. \cite{silver2016mastering} use Imitation Learning followed by Reinforcement Learning. Kai-Wei, et al. \cite{kai2015learning} use Monte Carlo estimates to calculate $Q^{*}(s,a)$, and train an apprentice $\pi$ to maximise $\sum_a \pi(a|s) Q^{*}(s,a)$. At each iteration after the first, the rollout policy is changed to a mixture of the most recent apprentice and the original expert. This too can be seen as blending an RL algorithm with Imitation Learning: it combines Policy Iteration and Imitation Learning.

Neither of these approaches is able to improve the original expert policy. They are useful when strong experts exist, but only at the beginning of training. In contrast, because \textsc{ExIt} creates stronger experts for itself, it is able to use experts throughout the training process.

AlphaGo Zero (AG0)\cite{silver2017mastering}, presents an independently developed version of ExIt, \footnote{Our original version, with only policy networks, was published before AG0 was published, but after its submission. Our value networks were developed before AG0 was published, and published after Silver et al.\cite{silver2017mastering}} and showed that it achieves state-of-the-art performance in Go. We include a detailed comparison of these closely related works in the appendix.

Unlike standard Imitation Learning methods, \textsc{ExIt} can be applied to the Reinforcement Learning problem: it makes no assumptions about the existence of a satisfactory expert. \textsc{ExIt} can be applied with no domain specific heuristics available, as we demonstrate in our experiment, where we used a general purpose search algorithm as our expert class.

\section{Conclusion}
\label{conclusion}

We have introduced a new Reinforcement Learning algorithm, Expert Iteration, motivated by the dual process theory of human thought. \textsc{ExIt} decomposes the Reinforcement Learning problem by separating the problems of generalisation and planning. Planning is performed on a case-by-case basis, and only once MCTS has found a significantly stronger plan is the resultant policy generalised. This allows for long-term planning, and results in faster learning and state-of-the-art final performance, particularly for challenging problems.

We show that this algorithm significantly outperforms a variant of the \textsc{REINFORCE} algorithm in learning to play the board game Hex. The resultant tree search algorithm beats MoHex 1.0, indicating competitiveness with state-of-the-art heuristic search methods, despite being trained tabula rasa.

\section*{Acknowledgements}

This work was supported by the Alan Turing Institute under the EPSRC grant EP/N510129/1 and by AWS Cloud Credits for Research. We thank Andrew Clarke for help with efficiently parallelising the generation of datasets, Alex Botev for assistance implementing the CNN, and Ryan Hayward for providing a tool to draw Hex positions. 

\nocite{young2016neurohex}
\nocite{Goldberg2013training}
\nocite{arpit2016normalization}
\nocite{clevert2015fast}
\bibliographystyle{unsrt}
\bibliography{bibliography}

\appendix

\newpage
\section{Comparison to AlphaGo Zero}
\label{ag0}

Silver et al. \cite{silver2017mastering} independently developed the \textsc{ExIt} algorithm, and applied it to Go. The result, AlphaGo Zero, outperformed the previous state-of-the art AlphaGo program while learning tabula rasa. Here we enumerate some of the similarities and differences between our implementations of the algorithm.

Both algorithms use tree policy targets for training the apprentice policy, however AlphaGo Zero uses Monte Carlo value estimates from the expert for the value network's training target, we use estimates from the apprentice. With approximately 100,000 times more computation during training than we used, the extra cost of such estimates is presumably outweighed by their greater accuracy. AlphaGo Zero includes all positions from all games in its dataset, whereas our positions are chosen to be independent samples from the exploration distribution.

In our implementation, we use the KL loss for both value and policy losses; AlphaGo Zero uses mean-square error for the value prediction. AlphaGo Zero also uses L2 regularisation of their network parameters, we use early stopping instead, and reinitialise neural network weights at each iteration before retraining on new data.

AlphaGo Zero uses a residual neural network with 79 layers, and significantly more units in each layer. Compared to a more standard CNN such as ours, they find significantly improved IL performance.

Our MCTS makes use of RAVE and rollouts, whereas AlphaGo Zero relies entirely on its value networks for evaluating moves. We also make use of warm-starting, progressing from a vanilla-MCTS expert to using only a policy network, and finally to using both policy and value networks. These warm starts are not essential for learning to play Hex, but they save substantially on computation early in training.

In AlphaGo Zero, before each expert improvement, they include verification that the new expert is indeed superior to the previous. They only change to a new expert if it defeats the current expert. Such tests would be prohibitively expensive compared to our training times, and we didn't find them to be necessary.

Unlike AlphaGo Zero, during dataset creation, we perform MCTS synchronously with no speed reduction, via the techniques described in appendix \ref{parallelisation}.

When creating training data, AlphaGo Zero adds Dirichlet noise to the prior policy at the root node of the search. This guarantees that all moves may be tried by the tree search, thus avoiding local optima. We do not use this trick, and our UCT formula includes a more usual exploration term, which guarantees that every move at the root will be attempted at least once, and so the apprentice policy will never completely disregard a move. We do not know whether noise at the root would help in Hex.

\section{Fast calculation of expert moves}
\label{parallelisation}

Because calculation of neural networks is faster when done in batches, and is performed on a GPU, most implementations of N-MCTS calculate their neural networks asynchronously: when a node is expanded, the position is added to a GPU calculation queue, but search continues. Once the queue length reaches the desired batch size $B$, the neural network policy can be calculated for the first $B$ states on the queue, and the information is added to the appropriate nodes.

Compared to waiting until the evaluation has taken place, this asynchronous neural net calculation substantially increases the rate at which MCTS iterations can take place: batching neural net calculations improves GPU throughput, and the CPU never sits idle waiting for evaluations. However, because search continues before evaluations are returned, suboptimal moves are made in the tree where prior information has not yet been calculated. 

In the \textsc{ExIt} setting, we can avoid asynchronous N-MCTS, CPU idle time and small calculation batches on our GPU. This is because we are creating a large dataset of N-MCTS moves, and can calculate multiple moves simultaneously. Suppose we have a set $P$ of positions to search from, and that $|P| > 2B$ . Each CPU thread gets a position $p_1$ from $P$, and continues search from that position until a NN evaluation is needed. It then saves the current search state (which can be expressed as a single pointer to the current tree node), and submits the necessary calculation to the GPU queue. It then moves on to another position $p_2$ from $P$, which isn't awaiting a neural net evaluation.

\section{Monte Carlo Tree Search Parameters and Rapid Action Value Estimation (RAVE)}
\label{RAVE}

RAVE is a technique for providing estimates of the values of moves in the search tree more rapidly in the early stages of exploring a state than is achieved with UCT alone. This is important, because the Monte Carlo value heuristic requires multiple samples to achieve a low variance estimate of the value, which is particularly problematic when there are many actions available.

A common property of many games is that a move that is strong at time $t_2$ is likely to have also been strong at time $t_1 < t_2$. For instance, in stone placing games such as Go and Hex, if claiming a cell is useful, it may also have been advantageous to claim it earlier. RAVE attempts to exploit this heuristic to harness estimates for many actions from a single rollout.

RAVE statistics $n_\mathrm{RAVE}(s)$, $n_\mathrm{RAVE}(s,a)$ and $r_\mathrm{RAVE}(s,a)$ are stored that correspond to the statistics used in normal UCT. After a simulation $s_1,a_1,s_2,a_2,...,s_T$, with result $R$, RAVE statistics are updated as follows:

\begin{equation*}
\begin{split}
n_\mathrm{RAVE}(s_{t_i},a_{t_j}) &:= n_\mathrm{RAVE}(s_{t_i},a_{t_j}) + 1\quad  \forall\ t_i < t_j \\
r_\mathrm{RAVE}(s_{t_i},a_{t_j}) &:= r_\mathrm{RAVE}(s_{t_i},a_{t_j}) + R\quad  \forall\ t_i < t_j \\
n_\mathrm{RAVE}(s_{t_i}) &:= \sum_a n_\mathrm{RAVE}(s_{t_i},a) \quad  \forall\ t_i
\end{split}
\end{equation*}

In other words, the statistics for state $s_{t_i}$ are updated for each action that came subsequently as if the action were taken first. This is also known as the \textit{all-moves-as-first} heuristic, and is applicable in any domain where actions can often be transposed.

To use the statistics a $\mathrm{UCT}_\mathrm{RAVE}$ is calculated, and averaged with the standard $\mathrm{UCT}$ into the tree policy to give $\mathrm{UCT}_\mathrm{U,RAVE}$, which then chooses the action. Specifically:
\begin{equation*}
\begin{split}
\mathrm{UCT}_\mathrm{RAVE}(s,a) =& \frac{r_\mathrm{RAVE}(s,a)}{n_\mathrm{RAVE}(s,a)} + c_b \sqrt{\frac{\log{n_\mathrm{RAVE}(s)}}{n_\mathrm{RAVE}(s,a)}} \\
\beta(s,a) =& \sqrt{\frac{c_\mathrm{RAVE}}{3n(s) + c_\mathrm{RAVE}}} \\
\mathrm{UCT}_\mathrm{U,RAVE} =& \beta(s,a)\mathrm{UCT}_\mathrm{RAVE}(s,a) \\&+ (1-\beta(s,a))\mathrm{UCT}(s,a)
\end{split}
\end{equation*}

The weight factor $\beta(s,a)$ trades the low variance values given by RAVE with the bias of that estimate. As the number of normal samples $n(s)$ increases, the weight given to the RAVE samples tends to 0. $c_\mathrm{RAVE}$ governs how quickly the RAVE values are down-weighted as the number of samples increases.

When using a policy network, the formulae are:
\begin{equation*}
\begin{split}
\mathrm{UCT}_{\mathrm{P-NN}}(s,a) =& \mathrm{UCT}(s,a) + w_a\frac{\hat{\pi}(a|s, \tau)}{n(s,a)+1} \\
UCT_\mathrm{P-NN,RAVE}(s, a) =& \beta(s,a)\mathrm{UCT}_\mathrm{RAVE}(s,a) \\&+(1-\beta(s,a))\mathrm{UCT}(s,a)\\&+ w_a\frac{\hat{\pi}(a|s, \tau)}{n(s,a)+1}
\end{split}
\end{equation*}

When using both policy and value estimates, the formulae are:

\begin{equation*}
\begin{split}
\mathrm{UCT}_{\mathrm{PV-NN}}(s,a) =& \mathrm{UCT}(s,a) + w_a\frac{\hat{\pi}(a|s, \tau)}{n(s,a)+1} + w_v \hat{Q}(s,a)\\
UCT_\mathrm{PV-NN,RAVE}(s, a) =& \beta(s,a)\mathrm{UCT}_\mathrm{RAVE}(s,a) \\&+(1-\beta(s,a))\mathrm{UCT}(s,a)\\&+ w_a\frac{\hat{\pi}(a|s, \tau)}{n(s,a)+1} + w_v \hat{Q}(s,a)
\end{split}
\end{equation*}

Where $\hat{Q}(s,a)$ is the backed up average of the network value estimates at the edge $s, a$.

\begin{table}[h]
	\caption{Monte Carlo Tree Search Parameters. Vanilla-MCTS refers to the parameters used in section \ref{ilhex}. N-MCTS parameters are for when only a policy network is used and when both policy and value networks are used.}
	\centering
	\begin{tabular}{llll}
		\toprule
		\cmidrule{1-4}
		Parameter     					& Vanilla-MCTS    	& N-MCTS (policy) 	& N-MCTS (policy \& value) \\
		\midrule
		Iterations 						& 10,000			& 10,000    	& 10,000 \\
		Exploration Constant $c_b$     	& 0.25			 	& 0.05     		& 0.05 \\
		$c_\mathrm{RAVE}$    			& 3000		       	& 3000 			& 3000 \\
		Expansion Threshold    			& 0			       	& 1			 	& 1 \\
		NN Weight $w_a$	    			& N/A       		& 100 			& 100 \\
		NN Output Softmax Temperature $\tau$	& N/A    	& 0.1 			& 0.1 \\
		Value Network Weight $w_v$      & N/A               & N/A           & 0.75
		\\		\bottomrule
	\end{tabular}
	
\end{table}

\section{Neural Network Architecture}
\label{nn_appendix}

\textbf{Input Features}. We use the same state representation as Young et al. \cite{young2016neurohex}: a two-dimensional state of $9 \times 9$ Hex board is extended to a 6 channel input. The 6 channels represent: black stone locations, white stone locations, black stones connected to the north edge, black stones connected to the south edge, white stones connected to the west edge and white stones connected to the east edge.

\begin{figure*}[ht]
	\begin{center}
		\includegraphics[width=\textwidth]{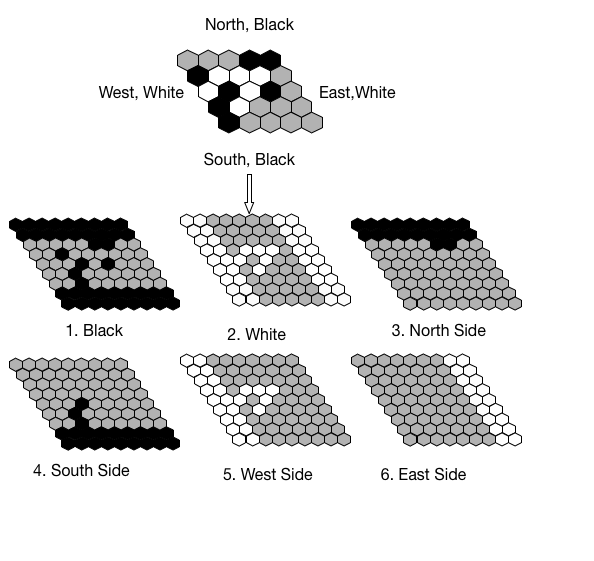}\\
		\caption{The 6-channel encoding used as input for our NN}
		\label{1to6channels}
	\end{center}
\end{figure*}

In line with Young et al. \cite{young2016neurohex}, to help with the use of convolutions up to the board edge, we also expand the board, adding two extra rows or columns to each side. On the extra cells thus created, we add dummy stones: along the North and South edges, black stones, along the East and West edges, White stones. In each corner of the padding, we `place both a black and a white stone'. The resultant encoding of the board is shown in figure \ref{1to6channels}.

Playing Hex on this expanded board, with the dummy stones providing connections in the same way as stones played by players, does not change the game, but it means that convolutions centred at the edge of the board have more meaningful input than would be provided by zero-padding these cells.

\textbf{Neural network architecture}. Our network has 13 convolution layers followed by 2 parallel fully connected softmax output layers.

The parallel softmax outputs represent the move probabilities if it is black to move, and the move probabilities if it is white to move. Before applying the softmax function, a mask is used to remove those moves which are invalid in the current position (i.e. those cells that already have a stone in them).

When also predicting state value, we add two more outputs, each outputting a scalar with a sigmoid non-linearity. This estimates the winning probability, depending on which player is next to move.

Because the board topology is a hexagonal grid, we use hexagonal filters for our convolutional layers. A $3\times3$ hexagonal filter centred on a cell covers that cell, and the 6 adjacent cells.

In convolution layers 1-8 and layer 12, the layer input is first zero padded and then convolved with 64 $3\times3$ hexagonal filters with stride 1. Thus the shape of the layer's output is the same as its input. Layers 9 and 10 do not pad their input, and layers 11 and 13 do not pad, and have $1\times1$ filters.

Our convolution layers use Exponential linear unit (ELU) \cite{clevert2015fast} nonlinearities. Different biases are used in each position for all convolution layers and normalisation propagation \cite{arpit2016normalization} is applied in layers 1-12. 

\begin{figure*}[h]
	\begin{center}
		\includegraphics[scale=0.5]{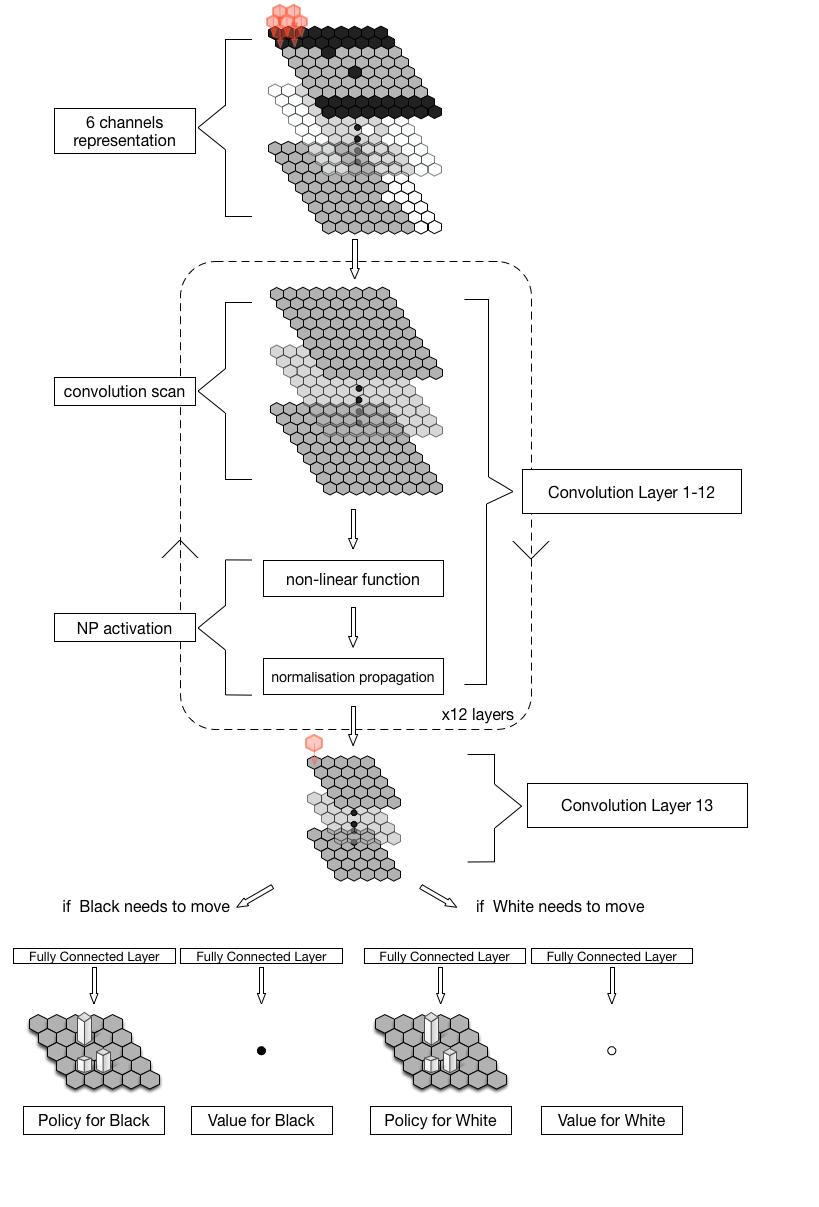}\\
		\caption{The NN architecture}
		\label{1net_np}
	\end{center}
\end{figure*}

This architecture is illustrated in figure \ref{1net_np}.

\textbf{Training details}. At each training step, a randomly selected mini batch of 250 samples is selected from the training data set and Adam \cite{kingma2014adam} is used as optimiser. We regularise the network with early stopping. The early stopping point is the first epoch after which the validation errors increase 3 times consecutively. An epoch is one iteration over each data point in the data set.

\clearpage

\section{Matches between \textsc{ExIt} and \textsc{MoHex}}

\textsc{ExIt} is clearly stronger than \textsc{MoHex} 1.0. \textsc{MoHex} 2.0 is ${\sim}250$ Elo stronger than \textsc{MoHex} 1.0 on $11\times11$, with the strength difference usually slightly lower on smaller boards. Although the most recent versions of \textsc{MoHex} are not available for comparison, we conclude that \textsc{ExIt} is competitive with state of the art for (scalable) heuristic search methods, but cannot say whether it exceeds it, particularly for larger 11x11 or 13x13 boards, which are more typical for tournament play. Future work should verify the scalability of the method, and compare directly to state-of-the-art methods.\footnote{Following out-of-date documentation on a \textsc{MoHex} repository, a previous version of this work stated that matches were played against \textsc{MoHex} 2.0. This was not the case; in fact there is no publicly published installation procedure for \textsc{MoHex} 2.0}

We present six games between \textsc{ExIt} (10,000 iterations per move) and \textsc{MoHex}\footnote{Source code from git://benzene.git.sourceforge.net/gitroot/benzene/benzene} (100,000 iterations per move), from a match of 162 games (consisting of one game each as black per legal opening move), of which \textsc{ExIt} won $59.3\%$. Here \textsc{MoHex} only uses its MCTS; in tournament play it uses DFPN solver in parallel to the MCTS. The games were chosen with `even' opening moves that don't give either player a large advantage, and to show some of the relative strengths and weaknesses of both algorithms. 

\begin{figure}[h!]
	\begin{center}
		\includegraphics[width=325pt]{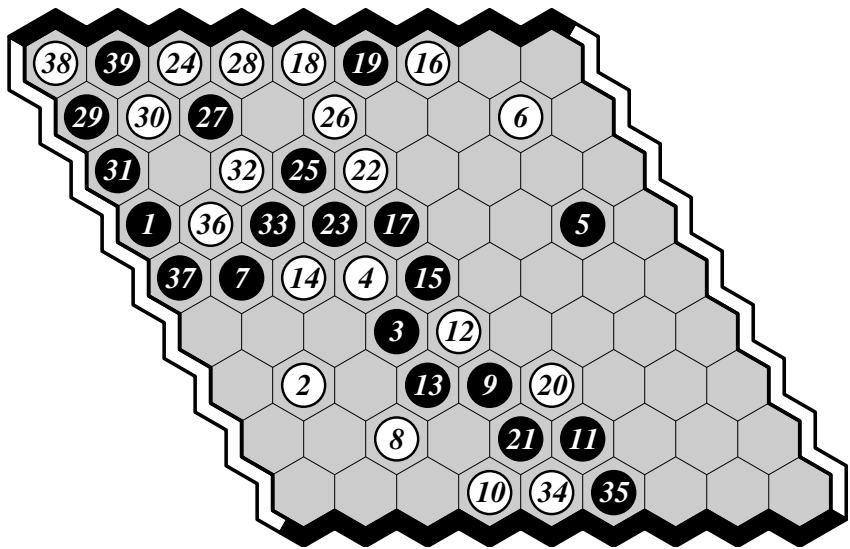}\\

		\caption{\textsc{ExIt} (black) vs \textsc{MoHex} (white)}
	\end{center}
\end{figure}

\begin{figure}[h!]
	\begin{center}
		\includegraphics[width=325pt]{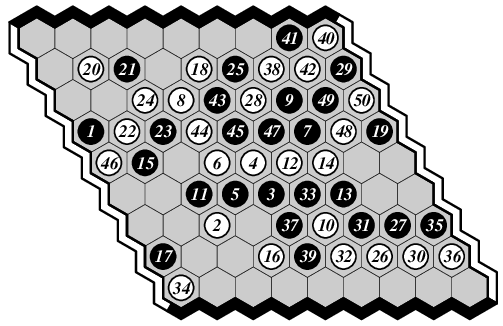}\\
		
		\caption{\textsc{MoHex} (black) vs \textsc{ExIt} (white)}
	\end{center}
\end{figure}

\begin{figure}[h!]
	\begin{center}
		\includegraphics[width=325pt]{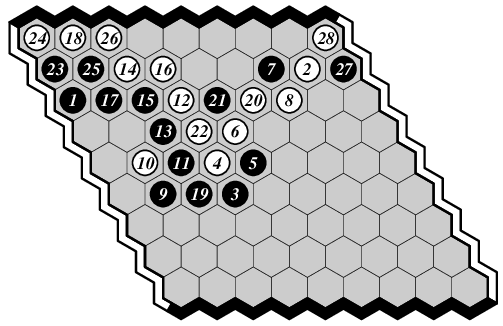}\\
		
		\caption{\textsc{ExIt} (black) vs \textsc{MoHex} (white)}
	\end{center}
\end{figure}

\begin{figure}[h!]
	\begin{center}
		\includegraphics[width=325pt]{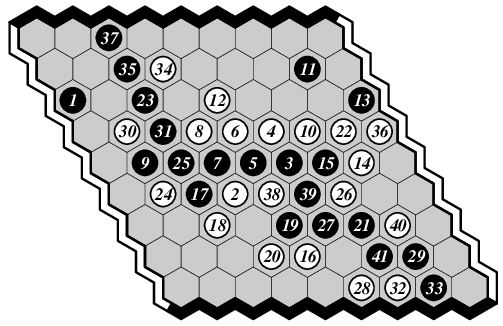}\\
		
		\caption{\textsc{MoHex} (black) vs \textsc{ExIt} (white)}
	\end{center}
\end{figure}

\begin{figure}[h!]
	\begin{center}
		\includegraphics[width=325pt]{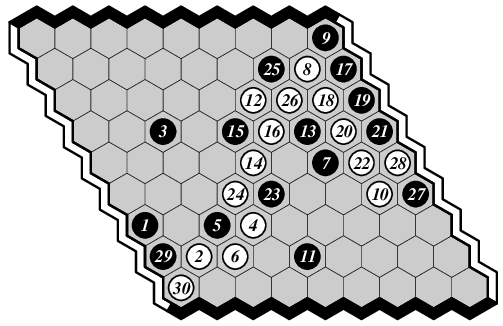}\\
		
		\caption{\textsc{ExIt} (black) vs \textsc{MoHex} (white)}
	\end{center}
\end{figure}

\begin{figure}[h!]
	\begin{center}
		\includegraphics[width=325pt]{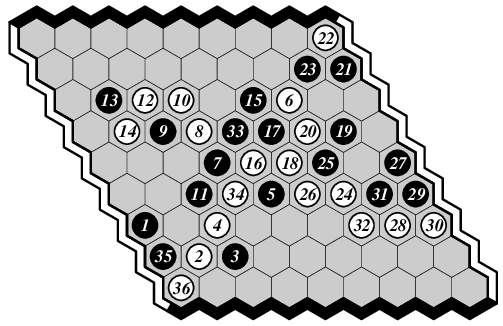}\\
		
		\caption{\textsc{MoHex} (black) vs \textsc{ExIt} (white)}
	\end{center}
\end{figure}

\end{document}